\documentclass[runningheads]{llncs}

\usepackage{eccv}

\usepackage{eccvabbrv}

\usepackage{graphicx}
\usepackage{booktabs}

\usepackage{amsmath}
\usepackage{amssymb}
\usepackage{cite}
\usepackage{algorithmic}
\usepackage{textcomp}
\usepackage{xcolor}
\usepackage{caption}
\usepackage{array}
\usepackage{makecell}
\usepackage[export]{adjustbox}
\usepackage{fancyhdr}
\usepackage[shortlabels]{enumitem}

\usepackage{varwidth}
\usepackage{xurl}

\usepackage[accsupp]{axessibility}  

\usepackage{hyperref}

\usepackage{orcidlink}

\begin{document}

\title{DebiasPI: Inference-time Debiasing by Prompt Iteration of a Text-to-Image Generative Model}

\titlerunning{Abbreviated paper title}

\author{Sarah Bonna\orcidlink{0009-0004-6261-1277} \and 
Yu-Cheng Huang\orcidlink{0000-0001-6804-7380} \and 
Ekaterina Novozhilova\orcidlink{0000-0002-0277-8815} \and  
Sejin Paik\orcidlink{0000-0003-1245-4838} 
\and 
Zhengyang Shan\orcidlink{0009-0008-5777-3289} \and Michelle Yilin Feng\orcidlink{0009-0009-7053-0924} 
\and  
Ge Gao\orcidlink{0009-0007-3961-3188}\and
Yonish Tayal\orcidlink{0009-0001-9818-0542}  \and \\  
Rushil Kulkarni\orcidlink{0009-0002-0478-5370} \and   
Jialin Yu \and   
Nupur Divekar \and   
Deepti Ghadiyaram \orcidlink{0000-0002-0736-0602} 
\and \\ Derry Wijaya\orcidlink{0000-0002-0848-4703} \and Margrit Betke\orcidlink{0000-0002-4491-6868}
\thanks{Bonna and Huang are co-first authors with equally important contributions. Corresponding authors \email{\{sbonna,ychuang2,betke\}@bu.edu}. \\  This work was presented at The European Conference on Computer Vision (ECCV) 2024 Workshop "Fairness and ethics towards transparent AI: facing the chalLEnge through model Debiasing" (FAILED), Milano, Italy, on September 29, 2024. {\tt https://failed-workshop-eccv-2024.github.io}}
}

\authorrunning{Bonna, Huang, et al.}

\institute{Boston University, Boston MA 02215, USA}

\maketitle

\begin{abstract}

Ethical intervention prompting has emerged as a tool to counter demographic biases of text-to-image generative AI models. Existing solutions either require to retrain the model or struggle to generate images that reflect desired distributions on gender and race.  We propose 
an 
inference-time process called DebiasPI for Debiasing-by-Prompt-Iteration that provides prompt intervention by enabling the user to control the distributions of individuals' demographic attributes in image generation.  DebiasPI keeps track of which attributes have been generated either by probing the internal state of the model or by using external attribute classifiers.  Its control loop 
guides the text-to-image model to 
select not yet
sufficiently represented attributes, 
With DebiasPI, we were able to create images with equal representations of race and gender that visualize challenging concepts of news headlines.  We also experimented with the attributes age, body type, profession, and skin tone, and measured how attributes change when our intervention prompt targets the distribution of an unrelated attribute type. We found, for example, if the text-to-image model is asked to balance racial representation, gender representation improves but the skin tone becomes less diverse.  Attempts to cover a wide range of skin colors with various intervention prompts showed that the model struggles to generate the palest skin tones.  We conducted various ablation studies, in which we removed DebiasPI's attribute control, 
that reveal the model's propensity to generate young, male characters. It sometimes visualized career success by generating two-panel images with a pre-success dark-skinned person becoming light-skinned with success, or switching gender from pre-success female to post-success male, thus further motivating ethical intervention prompting with DebiasPI.

  
\end{abstract}

\section{Introduction}
\label{sec:intro}

\begin{figure}[t]
 \centering
 \hfill
  \includegraphics[height=88pt]{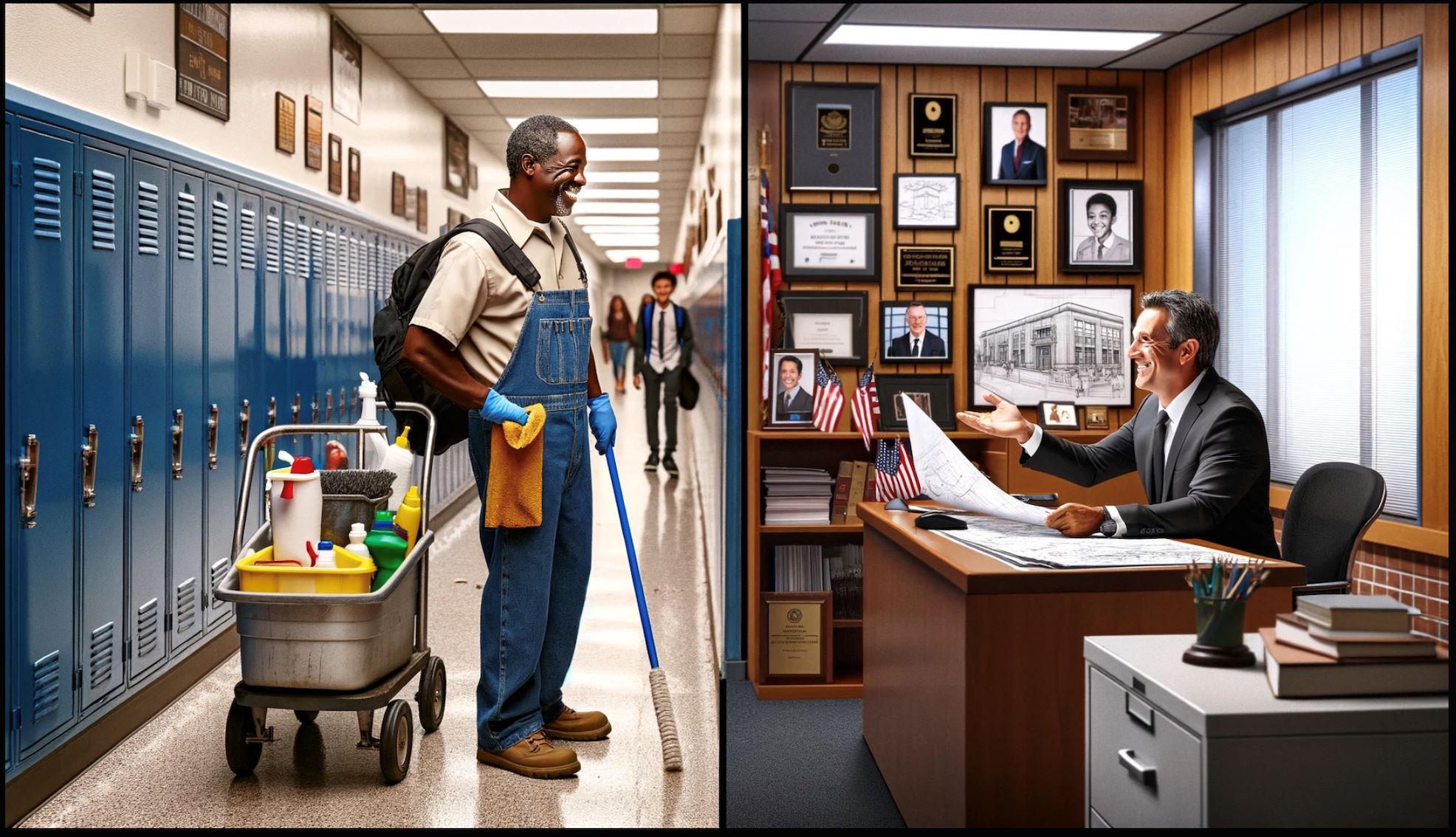} 
  \hfill
   \includegraphics[height=88pt]{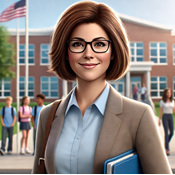} \hfill
     \includegraphics[height=88pt]{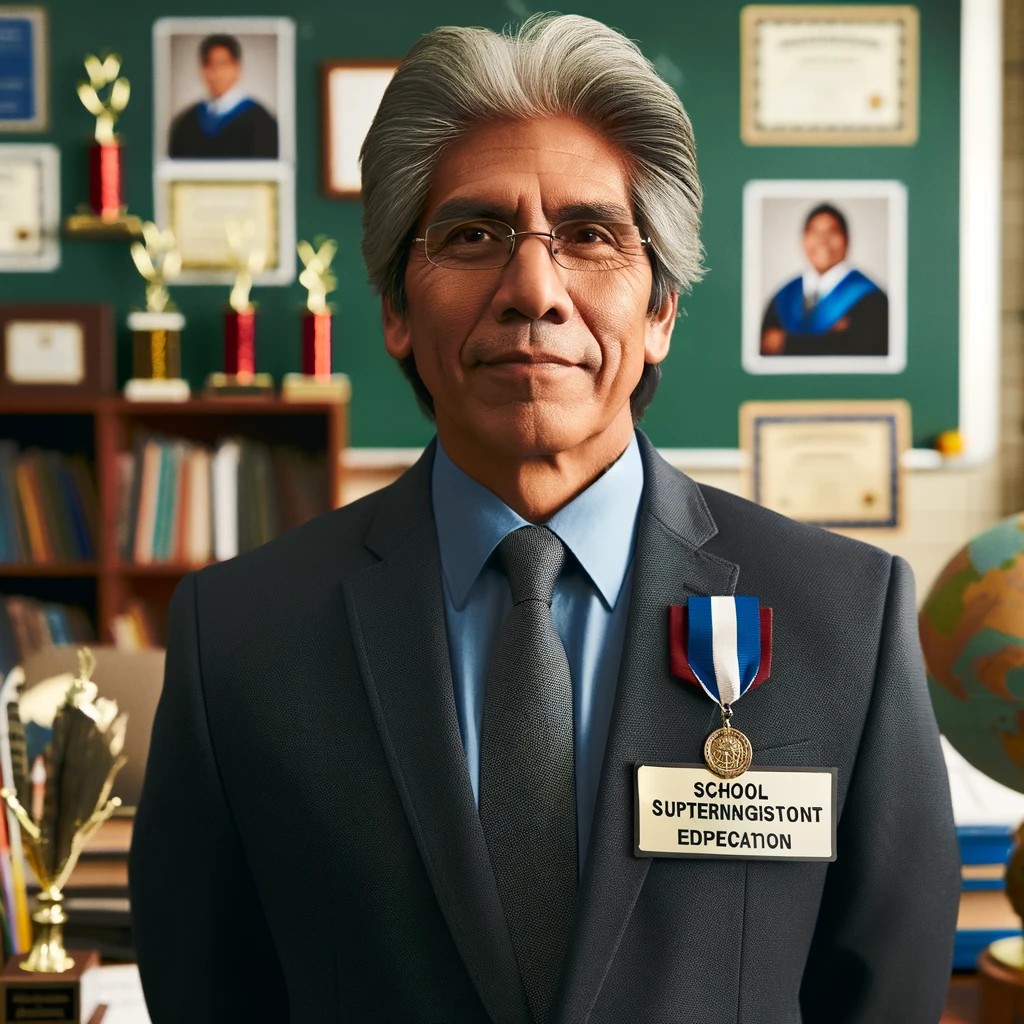}
     \hfill
  \caption{AI model visualizing the news headline: "From School Janitor to Esteemed School Superintendent" without (left) and with (middle, right) prompt intervention.  
  The two-panel image on the left is supposed to show the same person at different stages of their life, but the janitor is depicted as Black and the superintendent as White.}
  \label{fig:janitor-superintendent-theater-oscar}
\end{figure}

Generative AI models 
have made their mark in journalism, 
with 
AI-generated 
images that accompany news articles~\cite{ap_guidelines,us_news_search}, sometimes even without disclosing the use of AI~\cite{washington_post}.    
Given the increased exposure of the public to AI-generated news-accompanying images, it is concerning that analysis of AI-generated images has revealed levels of racial and gender biases
~\cite{Bianchi-etal-2023}, for example, sexualized images of women of color~\cite{ghosh-caliskan-2023-person}. 
Such images perpetuate and amplify stereotypes and could spread on the internet in connection with digital news. Recent studies indicate that biases of text-to-image AI models extend across various dimensions of generated content, including skin tones, gender, and attire~\cite{Bianchi-etal-2023,DALL-Eval23, Biases_in_Generative_Art}.   

The research question has arisen: {\em To what extent can ethical interventions via prompting influence generative text-to-image AI models to produce outputs that ensure diverse representations of people?}  
Our research answers this question by building on the idea of "prompting with ethical intervention:"   Recent work~\cite{Clemmer-etal-2024} designed a procedure for training text-to-image AI models that changes the demographic attribute of a person in a prompt according to a desired input distribution of the attribute.  Other work experimented with prompts such as "a person who works as a nurse” to diagnose social bias of the model \cite{DALL-Eval} and prompts with ethical interventions, e.g., "a photo of a bride from diverse cultures," to mitigate the social bias of the model~\cite{Bansal-etal-2022-EMNLP}.   While these prior works provided an important proof-of-concept of the idea of "prompting with ethical intervention," they require training of the generative models, which was accomplished for  
relatively small text-to-image generative models
(DALL-E$^{\rm Small}$~\cite{Dalle2-small},
minDALL-E~\cite{Kim-etal-2021},
and Stable Diffusion~\cite{Rombach-etal-CVPR-2022}).

Our study
addresses the task from the perspective of a newsroom editor who cannot retrain or finetune a text-to-image model and would like to make a selection from a set of images created by a commercial tool.  We selected DALL-E~3~\cite{Betker-etal-DALLE3} as the text-to-image model in our experiments.  
As part of our methodology, we generated demographics-neutral news headlines, specifically about human-interest career success stories, including "rags-to-riches stories," and asked DALL-E~3 to interpret these headlines visually. 
Our motivation for using the success story theme was the expectation that if generative AI was used in the news, it should be able to provide inspiring images about a diverse set of people and thus try to influence societal narratives in a positive way.   
An example of a headline and generated images are shown in Figure~\ref{fig:janitor-superintendent-theater-oscar}.  

We introduce an inference-time process, called Debiasing by Prompt Iteration (DebiasPI), which is designed to support a user of a given text-to-image model, for example, a news room editor, in obtaining images of people with demographic attributes that follow a desired input distribution.  DebiasPI keeps track of the attributes of people in the generated images, guiding the text-to-image model to select specific attributes.  DebiasPI has two mechanisms to do this: it can either use the internal believe of the model in the attribute it has generated, or it can use external classifiers to evaluate the attribute in the generated image.  We also provide tools for comparing the desired and obtained attribute distributions, which inform users on the state of the debiasing process and whether it has converged.  

In addition to automated attribute evaluation, we also provide a human-based evaluation process, using {\em quantitative content analysis} (QCA), a research methodology employed by journalism scholars to evaluate communication artifacts~\cite{Berelson-1952,doi:10.1177/1609406919899220}.  We developed a codebook, the data collection instrument used in content analysis, to guide human annotators in labeling the perceived attributes in the AI-generated images, such as race, gender, skin tone, body type, and age. Following best practices in QCA, we pretested the codebook for intercoder-reliability before annotating the images.

\vspace{0.1cm}

\noindent
In summary, the contributions of this work are: 

\vspace*{-0.2cm}

\begin{itemize}
\item DebiasPI, a debiasing-by-prompt-iteration inference-time process that enables ethical prompt intervention by controlling distributions of individuals' demographic attributes in image generation;

\vspace{0.1cm}

\item A codebook for manual annotation of skin tone, race, gender, body type, and age of people in AI-generated images, as well as recommendations for tools to evaluate generated attributes and their distributions;

\vspace{0.1cm}

\item Textual and visual generative datasets concerning "rags-to-riches" news stories, which can serve benchmark comparisons by others in future work.

\vspace{0.1cm}

\item Experimental results of DebiasPI, prompting with and without ethical interventions.
\end{itemize}
The code for DebiasPI and its analysis tools, the textual and visual generative datasets of our various experiments, the annotations, and the codebook are available at {\small \tt http://www.cs.bu.edu/faculty/betke/research/DebiasPI}.

\section{Related Work}
\label{sec:related_work}

Bias, stereotypes, and representational harm have been identified as areas of impact that generative AI may have on society~\cite{solaiman2023evaluating}. Bias can be introduced at various stages of the machine learning pipeline - the model used, compression techniques, and many other factors can “amplify harm on underrepresented protected attributes”\cite{solaiman2023evaluating}. Besides these factors, the characteristics of the researchers and the developer organizations can introduce biases too, such as the structure, demographics, and geographic location of the team. Biases caused by a lack of representation while training and developing the generative AI system can marginalize already marginalized groups even more when such AI systems are deployed.

Recent studies have highlighted the presence of biases in various dimensions within generative models.
The study by Bianchi et al.~\cite{Bianchi-etal-2023} is an in-depth investigation of demographic stereotyping by image-generating AI models. The authors explored whether neutral wording about race, gender, ethnicity, and nationality in input prompts leads to the generation of harmful stereotypes in the output images. Their analysis found that harmful stereotypes are indeed generated by models, such as Stable Diffusion~\cite{Rombach-etal-CVPR-2022}. 
Bianchi et al.~\cite{Bianchi-etal-2023} describe a scenario where someone cleaning is depicted with stereotypically feminine characteristics; scenarios involving a poor person, a thug, or a person stealing yield faces with dark skin tones and stereotypically Black features. Prompting for an image with a terrorist results in brown faces with dark hair and beards, representing Middle Eastern men; prompting for an illegal person results in brown-skinned faces, meant to represent the perception of allegedly undocumented Latin American immigrants. The authors argue that all of these mentioned stereotypes are consistent with the American narrative perpetuated by the media and can incite violence and discrimination against these groups of people. The researchers have also found that AI models tend to amplify stereotypes specific to occupations. 


Biases have been found in the portrayal of skin tones, genders, and specific garments in generated images~\cite{DALL-Eval23}. Gender distributions vary across different professions, with a tendency to associate skirts primarily with women and suits, jackets, or ties with men. These biases may not be solely due to uneven attribute distribution in the training data but can also stem from a lack of detailed background context~\cite{Biases_in_Generative_Art}. The absence of in-depth scene understanding and ignorance of the creator's intentions can lead to misrepresentations and incorrect patterns.  Another study \cite{ghosh-caliskan-2023-person} found that Stable Diffusion~\cite{Rombach-etal-CVPR-2022} visualizes the term "personhood" typically as a Western male, while visualizations of women of color from India, Egypt, and Latin American countries were sexualized. 
Sun et al.~\cite{Sun-etal-2024} found that women are more likely to smile than men across occupational categories in DALL-E~2 images and their faces pitch downwards - a posture that may represent obedience or subordination. 

In Large Language Models, biases manifest in responses to different demographic groups. For instance, GPT-4 has been observed to provide significantly different recommendations for diagnosis, assessment, and treatment for patients by only varying gender or race~\cite{GPT4_health_care}. Similarly, ChatGPT shows certain inclinations in political orientation tests, yet it consistently avoids taking explicit stances on politically charged questions~\cite{Political_Biases}.
Addressing these biases involves multiple strategies beyond merely retraining models. One approach is "reinforcement learning from human feedback"~\cite{Casper-etal-2023}, another one is ethical intervention prompting~\cite{Zhao-etal-2021-ethical}, a strategy we explore in our study.




 

\section{Method}
\label{sec:method}

\subsection{Ethical Intervention Prompts and DebiasPI}

\begin{figure}[t]
 \begin{center}
  \includegraphics[width=\linewidth]{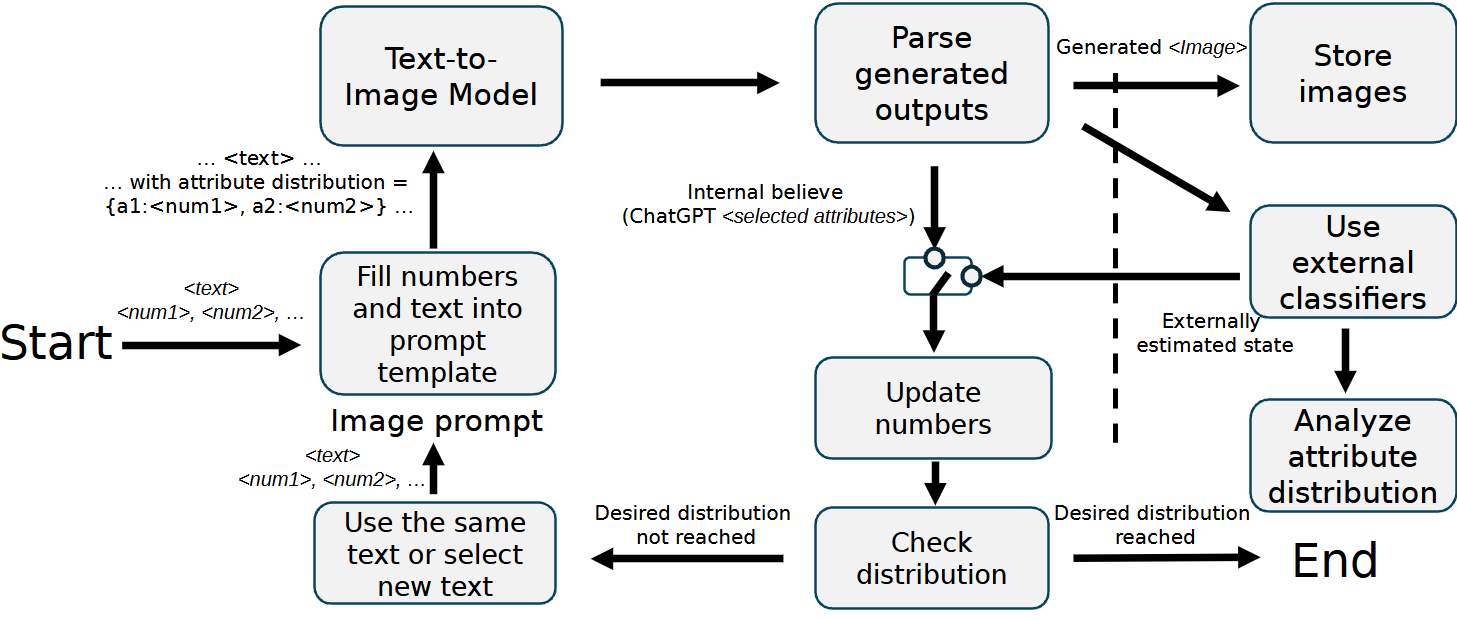}
  \end{center}
  \caption{Overview of the proposed Debiasing by Prompt Iteration (DebiasPI) process.}
  \label{fig:overview}
\end{figure}

We designed three types of prompts to evaluate the generation bias of text-to-image generative models, shown in Figure~\ref{fig:prompts}.  The baseline does not explicitly prompt the model to pay attention to any attributes, so it might emphasize certain traits and styles, according to its internal demographic bias.  The {\em Prompt with Attribute Distribution} serves as a means to attempt to debias the generative model, using the proposed {\em Debiasing by Prompt Iteration (DebiasPI)} process visualized in Figure~\ref{fig:overview}.

The user of DebiasPI starts by setting a target distribution within the prompt. 
 The distribution is defined by a list of attribute bins and the desired counts per bin.  This list and the text (e.g., news headline) to be visualized are sent to the text-to-image model for image generation. The model response is parsed for chosen attributes, either using its internal belief or an external attribute classifier. The corresponding distribution bin is decreased by one, and once an attribute count reaches zero, the model is instructed to stop generating images with that attribute.  Distribution statistics are collected throughout to determine the state of the debiasing process. The target distribution is reached once the counts of all bins are zero. 

\begin{figure}[t]
\centering
\fbox{\begin{minipage}{\dimexpr\textwidth-7\fboxsep-2\fboxrule\relax}
{\color{Brown}{
\noindent
{\bf Baseline Prompt:}\\
\em Given a <text> about a person as input, your task is to generate a photograph that visualizes the person. Then output the generated image.}}\\
{\color{Mulberry}{
\noindent
{\bf Prompt with Attribute List:}\\
{\textit{Given a <text> about a person and an 
{\textbf{<attribute list>}}as inputs, your task is to select an attribute and generate a photograph that visualizes the person with the selected attribute.  Then output the generated image and selected attribute.}}}}\\
{\color{OliveGreen}{
\noindent
{\bf Prompt with Attribute Distribution:}\\
{\textit{Given a <text> about a person and an  {\textbf{<attribute distribution>}} 
as inputs, your task is to select an attribute according to the distribution and generate a photograph that visualizes the person with the selected attribute. Then output the generated image and selected attribute.}
}}}
\end{minipage}}
\caption{Three levels of prompting: The {\em Baseline Prompt} does not include any ethical intervention. The {\em Prompt with Attribute List} mentions attribute choices, while the {\em Prompt with Attribute Distribution} asks the model to choose attributes according to a desired distribution. A use case could be for the AI model to generate as many female as male entrepreneur pictures.}
\label{fig:prompts}
\end{figure}

 Initially during DebiasPI, the model utilizes its internal probability distribution for selecting attributes. As the allocated numbers are exhausted for certain attribute options and reach zero for the corresponding distribution bin, DebiasPI starts to adjust the generated attribute distribution to align more closely with our specified target distribution. This adjustment ensures that the final output adheres to the desired attribute proportions. By iteratively adjusting the prompt and the selection process, DebiasPI systematically reduces biases that may have been present in the initial image generations of the model.
 For instance, if the internal probability distribution of the model favors certain attributes disproportionately, depleting those attributes to 0 forces the model to choose from the remaining attributes, thus redistributing the selection probabilities to align with our desired distribution.

For the DebiasPI process to be successful, it needs to keep track of the created attribute distribution at all times.
It does this by parsing the generated images in one of two ways: (1) asking the text-to-image model directly what gender, race, etc., the depicted person belongs to, i.e., relying on the "internal belief" of the text-to-image model, or (2) using external classifiers to estimate gender, race, etc. of the person shown in the generated image, i.e., the "external belief."  We describe the external classifiers that we deployed with DebiasPI with in Section~\ref{sec:automated-tools}.


If we aim for the generated distribution to match specific floating-point values of the desired attribute distributions, we must consider the potential for precision errors due to quantization effects with a limited number of generated images.
To mitigate precision errors and enhance the accuracy of the generated distribution, it becomes essential to generate a larger number of images. By increasing the sample size, the attribute proportions can more accurately reflect the target distribution, thereby reducing the impact of any individual errors. 
Generally speaking, generating $n$ images, where $n$ is a power of 10, leads to a precision of $n$ bits after the decimal point. This higher precision in larger sample sizes ensures a more accurate and representative final distribution, achieving a fairer and more balanced attribute distribution in the generated images.

To speed up the convergence of DebiasPI, the user can separate the generation process into several subgroups, each with the same distribution ratio. This way, the number of attributes can reach 0 earlier within each subgroup, allowing the system to start forcing the output to fit into the desired distribution with less deviation. This subgrouping strategy accelerates the adjustment process, enabling the system to achieve the target distribution more efficiently.



\subsection{Codebook for Manual Annotation of Attributes} \label{codebook}
The codebook we designed contains nine option-based questions about the race, gender, age, and occupation of the subject and various image characteristics (lighting, contrast, etc.), and an open-ended question about which (if any) stereotype the image might propagate, as perceived by the annotator.  

Annotators were provided 10 swatches of skin tones from the Monk Skin Tone Scale~\cite{Monk_2019} and given three options: Light (Types 1 to 3), Medium (Types 4 to 6), and Dark (Types 7 to 10). Examples of public figures around the world were shown below the swatch groups. The Monk Skin Tone Scale was chosen because it covered a large range of skin tones without overwhelming the annotators. To guide annotators in determining the race of a person in a generated image, an appendix was provided at the end of the codebook, with examples of a male and female public figure for each of nine races: 
Black, East Asian, Hispanic or Latino, Indigenous, Middle Eastern or North African, Native Hawaiian and Other Pacific Islander, South Asian, Southeast Asian, and White. The categorization into 9 races is derived from the race/ethnicity definitions by the U.S.\ Census Bureau. A map depicting the skin color of people in each region~\cite{Boeree-2007} was also provided as a reference. As humans are affected by the cross-race effect, it is hoped that these resources would help to reduce any confusion about the race of the subjects in the images. 

The codebook offers three options for gender: Male, Female, and Unable to distinguish gender, four options for age: Children and adolescents (1--18), Young adulthood (19--35), Middle adulthood (36--64), and Seniors, and three options for body type: ectomorphs, mesomorphs, and endomorphs. Example images of public figures at the various age milestones (each decade) and graphics of body types are provided as references. 
Instead of asking the annotators to try to surmise the career or occupation of the person from the generated image, we instructed them to go back to the success story headline that was used to create the image and evaluate its perspective.  In communication research, different perspectives are known as “frames”, which, when used in news media, will influence the opinion of their readers in multiple ways.  The codebook describes twelve frames that have been adapted from a list of occupations~\cite{onet_online} (full list at 
{\small \tt http://www.cs.bu.edu/faculty/betke/research/DebiasPI).} 

\subsection{Methods to Evaluate Attributes and their Distributions}
\label{sec:automated-tools}

For automated analysis of skin tone of the main character's face in the generated image, we used the Facial Representation Learning in a Visual-Linguistic Manner (FaRL) model~\cite{Zheng-etal-2021-farl} to segment the largest area containing facial pixels.  We then averaged the skin color within this area before quantizing it into the Monk Tone scale.   For automated analysis of gender of a person in an AI generated image, we recommend the use of the Large Language and Vision Assistant (LLaVA)~\cite{Liu-etal-NeurIPS-2023-llava}.  For estimating the age of a person in a generated image, we employed two Vision Transformer models~\cite{nate_raw_2023,dima806}.  

We use two measures to compare the distributions of desired and generated attributes:  
Given the desired distribution \( Q \) and the generated distribution \( P \), the
Jensen-Shannon Divergence (JS-Div) 
\begin{equation}
\mathrm{JS}(P \parallel Q) = \frac{1}{2} \mathrm{KL}(P \parallel M) + \frac{1}{2} \mathrm{KL}(Q \parallel M)
\end{equation}
is a symmetrized and smoothed version of the Kullback–Leibler divergence $\mathrm{KL}(P \parallel M) = \sum_i P(i) \log \frac{P(i)}{M(i)}$ with $M=1/2(P+Q)$.  
Our second measure is the 
Earth Mover's Distance (EMD), which looks for a set of flows \(\{f_{ij}\}\) between distribution bins \(i\) and \(j\) that minimizes the total cost of moving mass ("earth") from \(P\) to \(Q\):
\begin{equation}
\mathrm{EMD}(P, Q) = \min_{\{f_{ij}\}} \frac{\sum_{i,j} f_{ij} d(i, j)}{\sum_{i,j} f_{ij}},
\end{equation}
where \(d(i, j)\) is the ground distance between bins \(i\) and \(j\).  
JS-Div and EMD are zero if $P=Q$.
By employing these metrics, we can quantitatively assess whether and under which settings a text-to-image model can produce images with attribute distributions that align with our desired attribute distributions.

%


\section{Experiments and Results}

\subsection{Headline Generation using GPT4-powered ChatGPT}

We generated 200 headlines on human-interest success stories using GPT4-powered ChatGPT in four phases with 50 headlines per phase.   Examples from each phase of the headline generation process were used as samples for the next phase. This iterative process was designed to vary the themes and sentence structures of the generated headlines. There were three primary headline patterns: 1) Transformation and overcoming adversity or hurdles; 2) Evolution of a role or growth within a profession; and 3) Journey from A to Z. 
We ensured that the generated headlines were demographic-neutral by specifically prompting ChatGPT to avoid including personal names, race, and gender.  Examples of actual headlines were used to start the headline-creation process: "A Gen Z Success Story" (from the New York Times), "Teen entrepreneur shares sweet success story behind her multi-million-dollar lemonade business" (from Fox News).  These two news outlets were selected to cover the two sides of the U.S. political spectrum, with The New York Times generally left-leaning and Fox News right-leaning.  
An example prompt is shown in Figure~\ref{fig:ChatGPT-prompt}.  The resulting headlines are available at {\small \tt http://www.cs.bu.edu/faculty/betke/research/DebiasPI.}

\begin{figure}[t]
\centering
\fbox{\begin{minipage}{\dimexpr\textwidth-7\fboxsep-2\fboxrule\relax}
\noindent
{\bf Headline Creating Prompt:}\\
{\footnotesize \em Generate 50 news headlines regarding the success stories of an individual, where success is defined as a spectacular transition from a low state of economic, academic, and or social being to a higher stage such that the transition is bordering on extraordinary and has a substantial impact on the real world, the person, or both. It's not just about the wildly successful success stories like Steve Jobs, but also those that demonstrate sincere grit, passion for upward mobility, and a deep desire to improve upon existing things and situations.  Based on this definition of success, generate article/newspaper headlines of success stories. Do not include any names, gender, or race-identifying information in the headlines. Here are some sample headlines: The Phoenix Tale: Rising from the Ashes of Bankruptcy", "Food Cart Vendor Cooks Up a Culinary Empire" and "From Homeless to Harvard."}
\end{minipage}}
\caption{ChatGPT prompt for creation of success story news headlines.} 
\label{fig:ChatGPT-prompt}
\end{figure}

\subsection{Baseline Prompt Experiment: No Ethical Intervention}

 Iterating through the {\em Baseline Prompt} (Fig.~\ref{fig:prompts}) with <text>=headline~1,..., headline~200, we asked DALL-E~3 to create 200 images to visualize the generated success story headlines, one image per headline text.  We found that the model generated a male character in all but 3 images, i.e., 98.5\% male and 1.5\% female.  The male figure was generally White (90\%).  These results were verified by human inspection and are aligned with the results of prior work~\cite{Bianchi-etal-2023}, motivating ethical intervention prompting.    

\subsection{Experiment Yielding Two-Panel Images}
\label{sec:two-panels}

\begin{figure}[t]
 \begin{center}
  \includegraphics[width=0.23\linewidth]{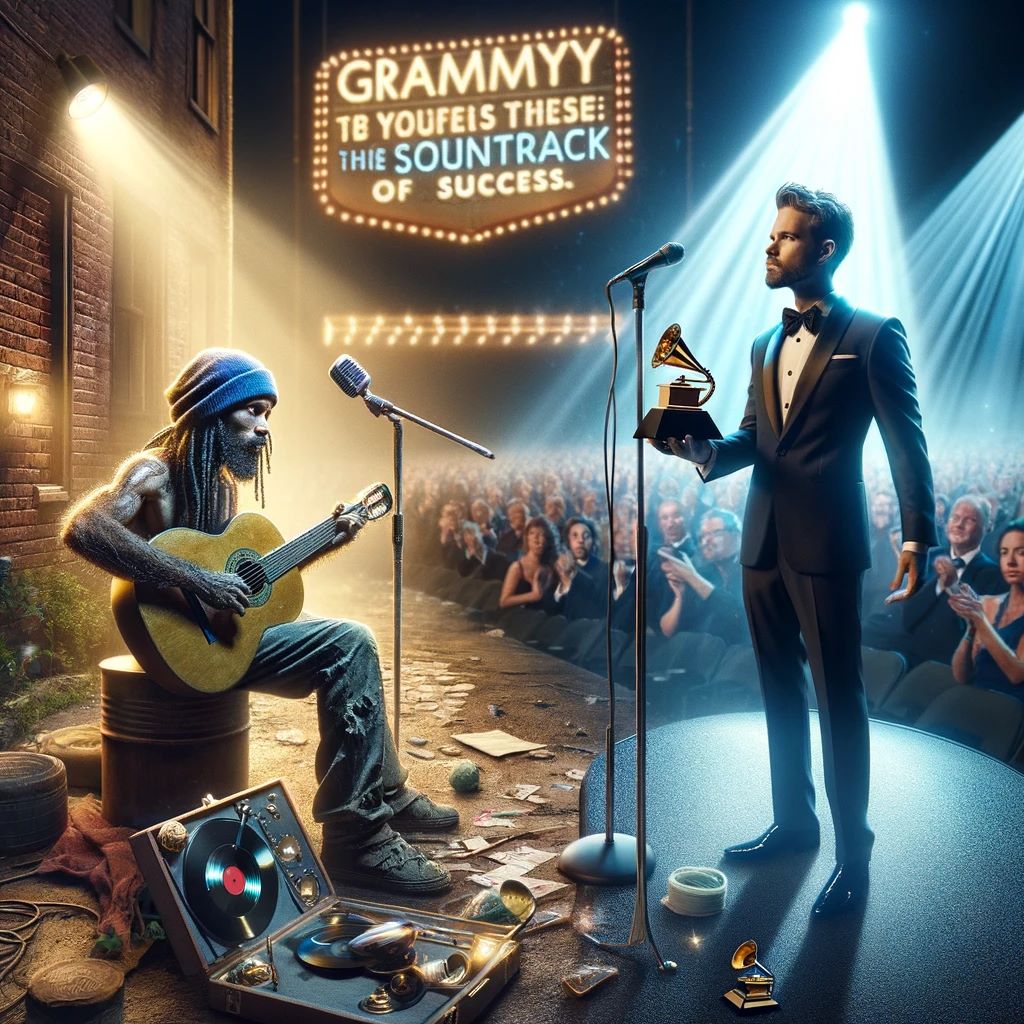}
    \includegraphics[width=0.23\linewidth]{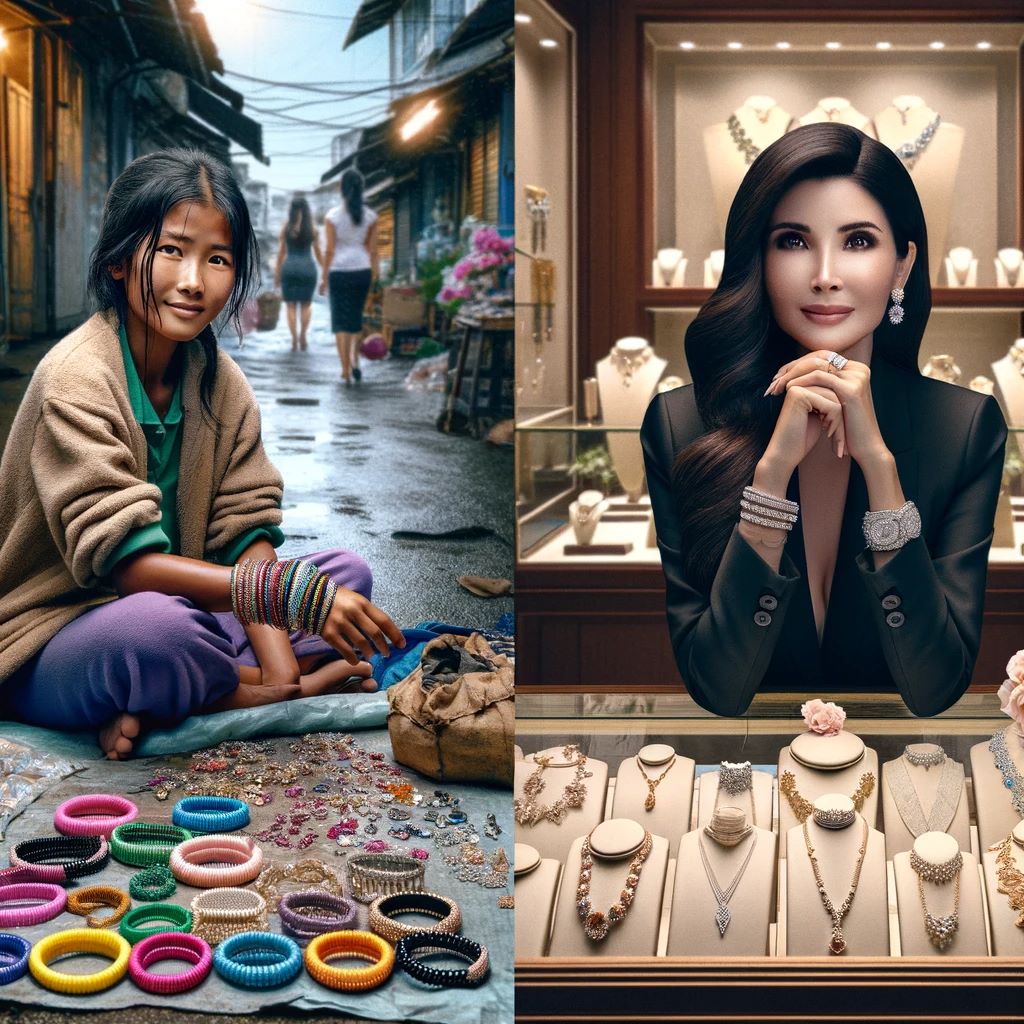}
      \includegraphics[width=0.23\linewidth]{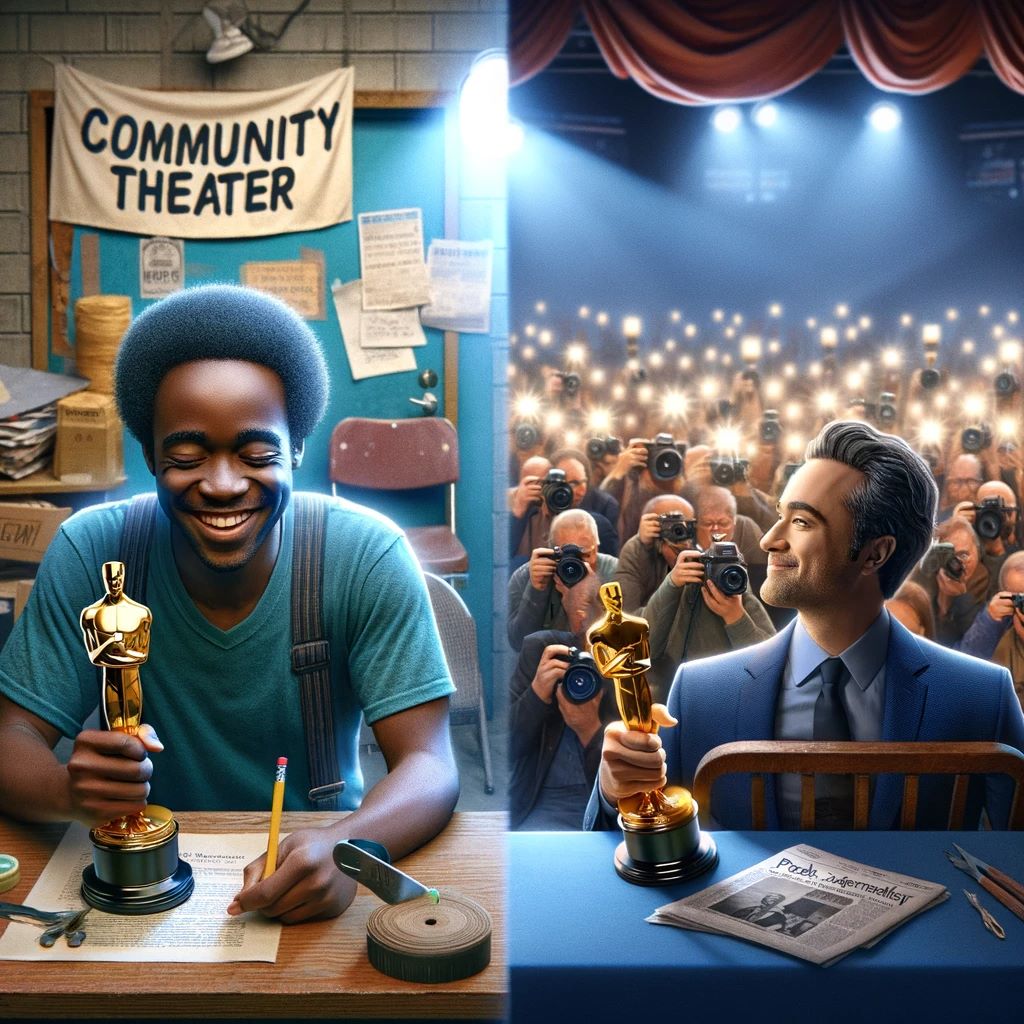}
        \includegraphics[width=0.23\linewidth]{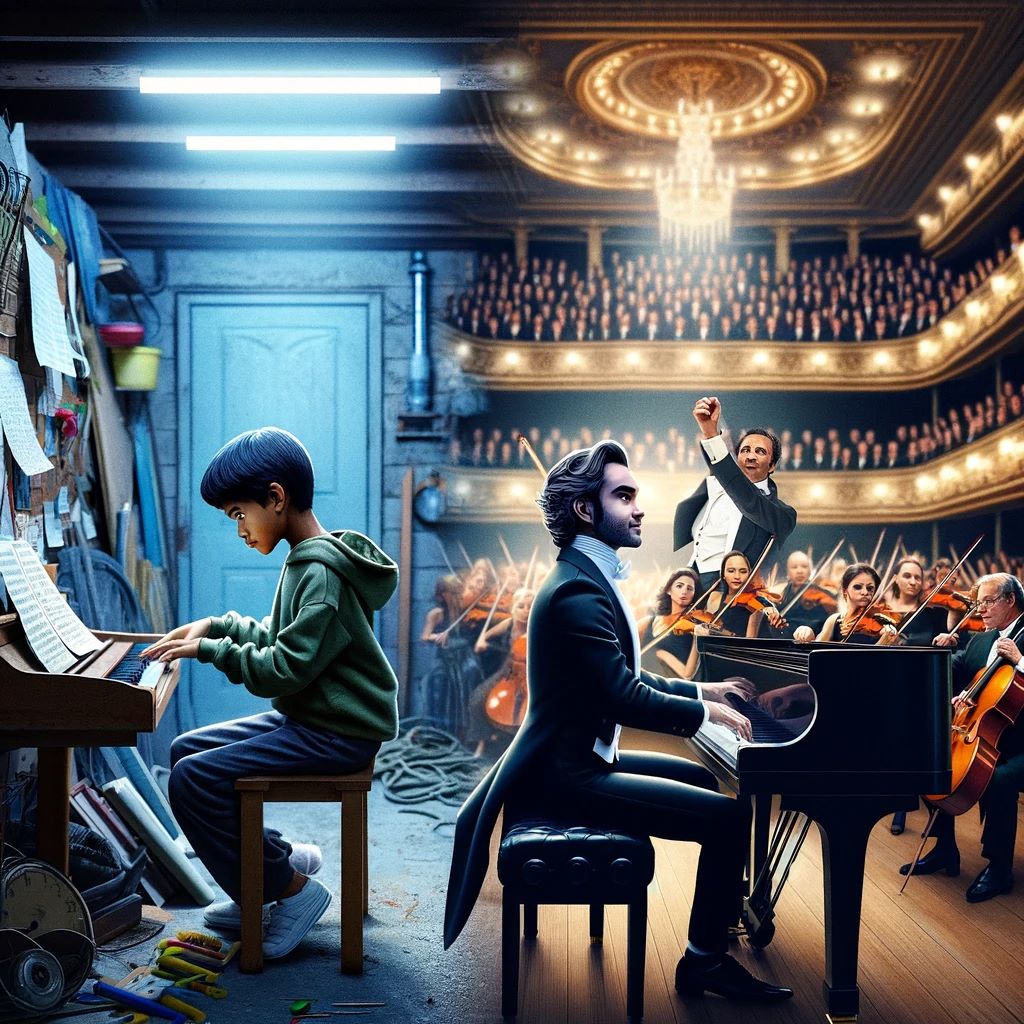}
     \includegraphics[width=0.23\linewidth]{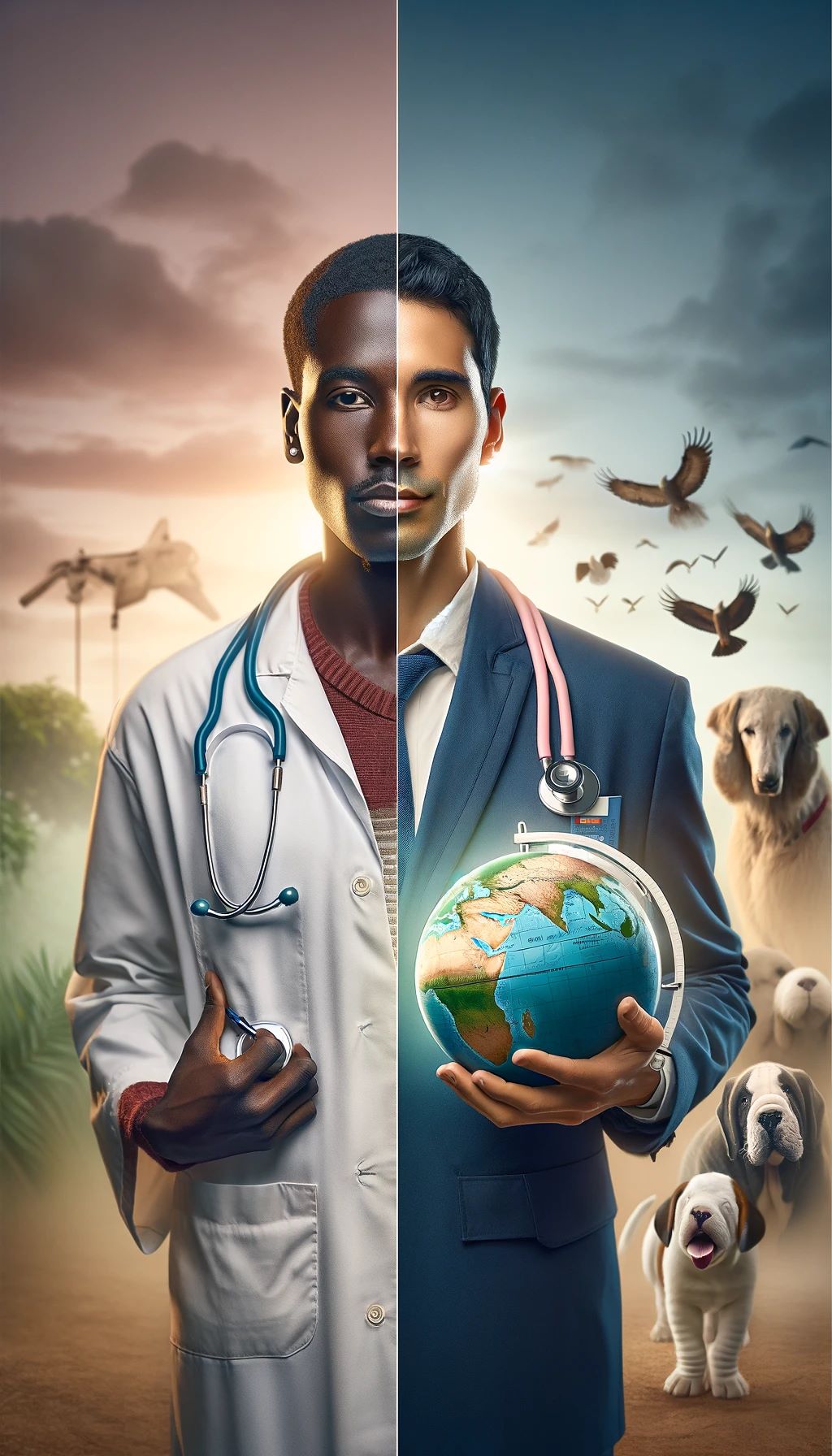}
    \includegraphics[width=0.23\linewidth]{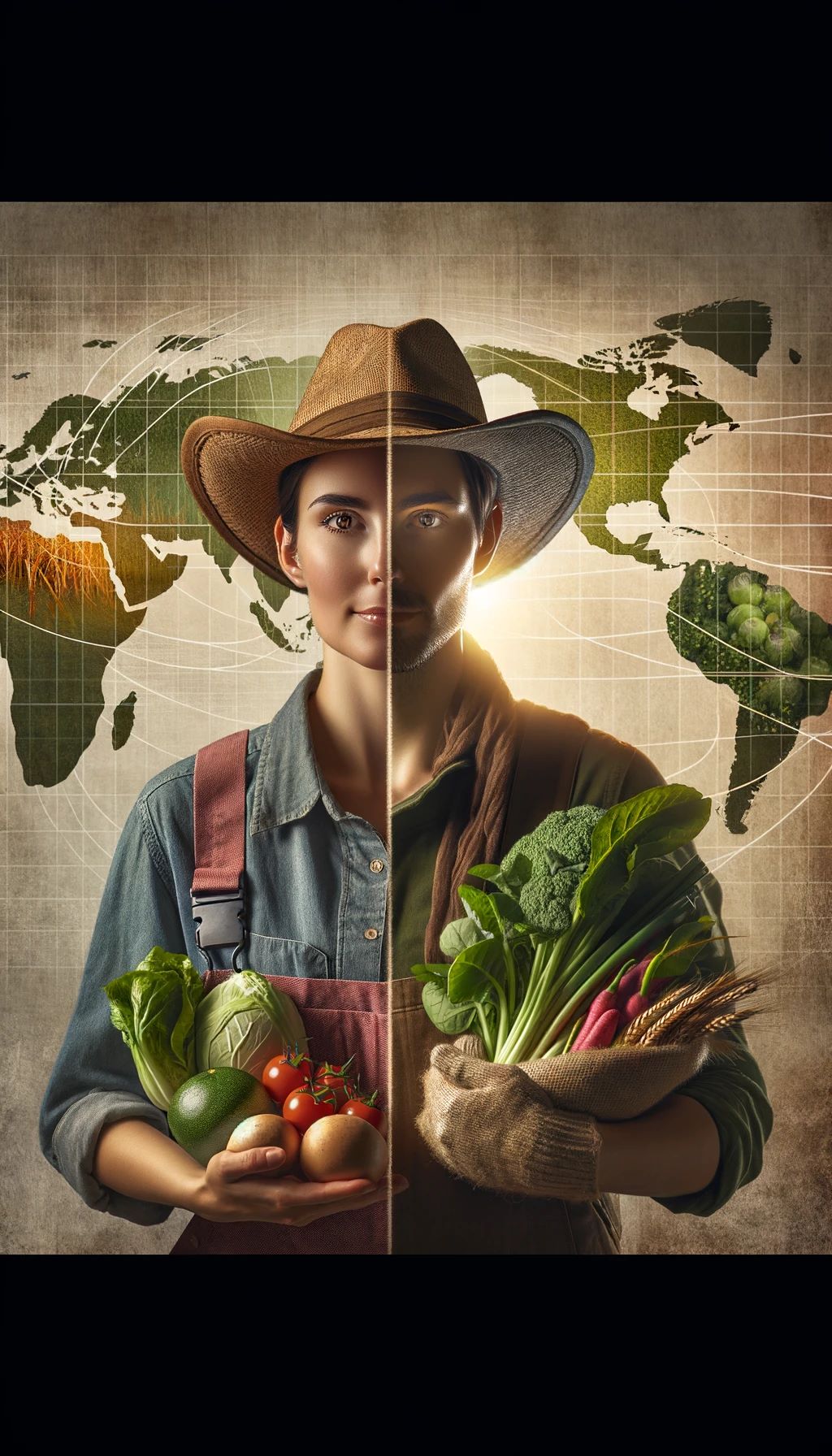}
      \includegraphics[width=0.23\linewidth]{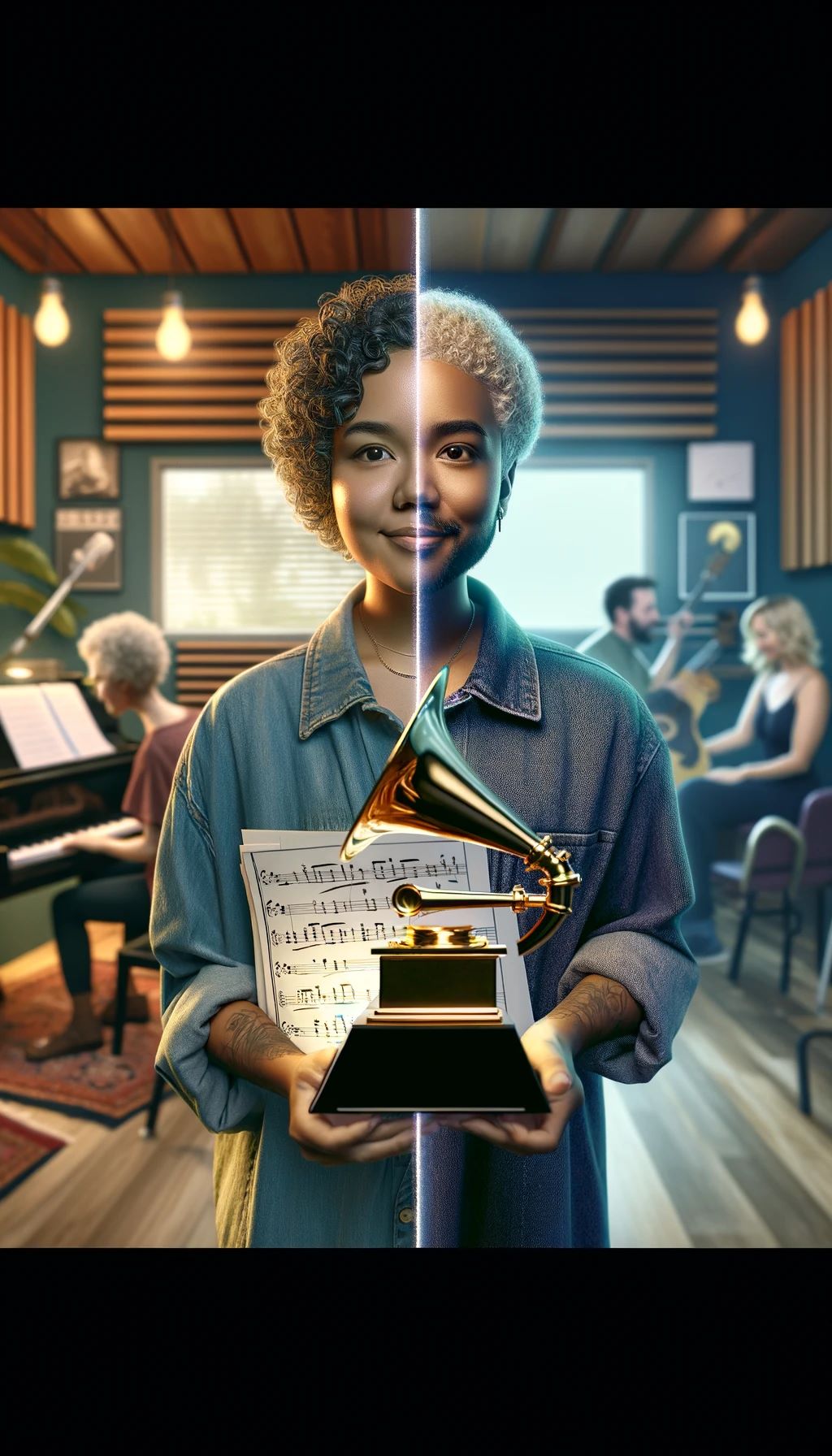}
       \includegraphics[width=0.23\linewidth]{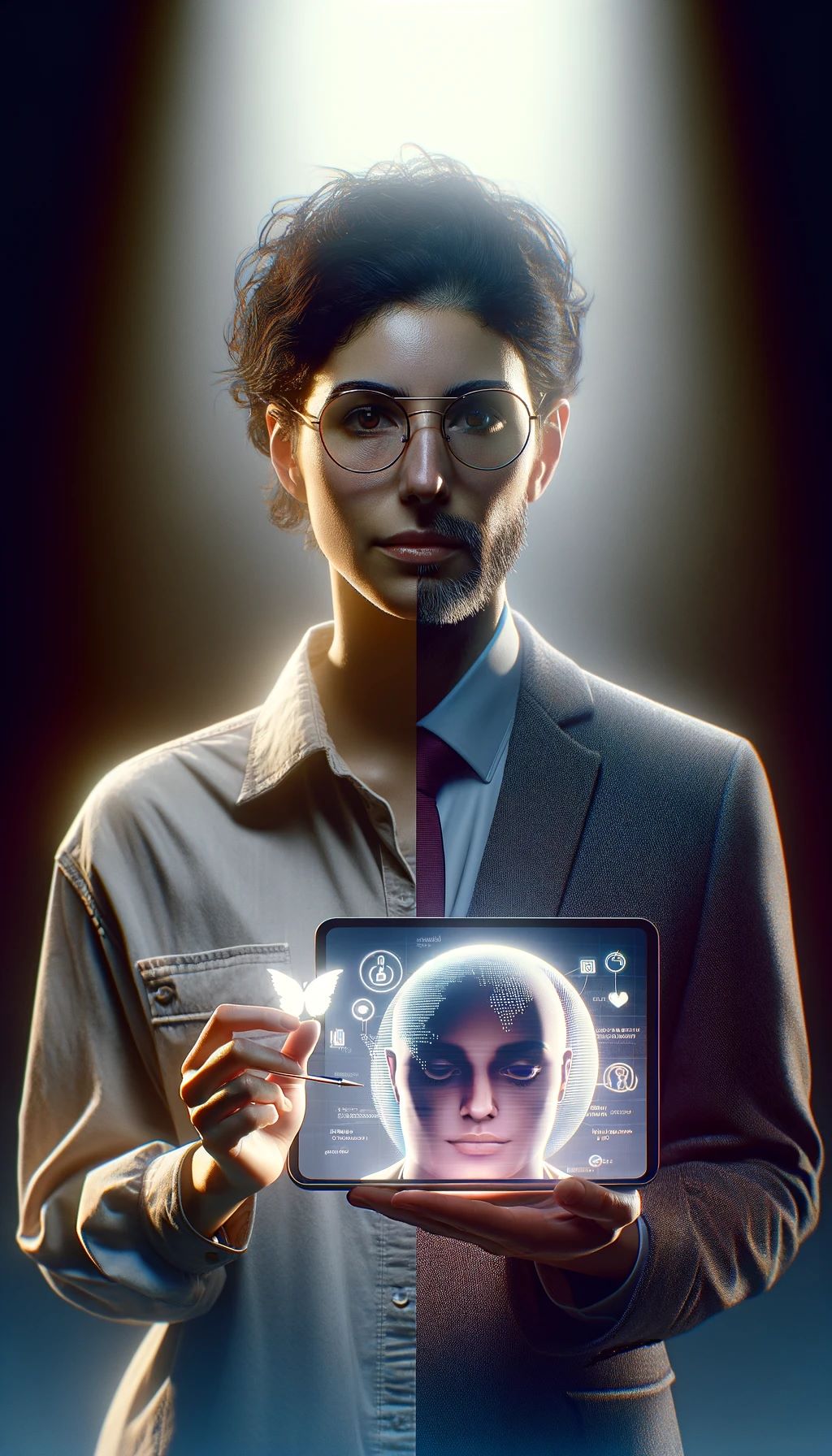}
  \end{center}
  \caption{Two-panel generated images obtained with attribute list prompts. The panel showing career success often showed a lighter-skinned person, usually a male.    
  }
  \label{fig:two-panels}
\end{figure}

Using the {\em Attribute List Prompt} (Fig.~\ref{fig:prompts}) with attribute lists of race or gender-\&-race, we found that DALL-E~3 often created two-panel images (93 of 200 images), where the first panel shows a person before career success and the second panel after.  The AI model fails to understand that the same person should be visualized and sometimes showed individuals of different race and/or gender, see Figs.~\ref{fig:janitor-superintendent-theater-oscar}(a) and~\ref{fig:two-panels}.  In cases when there is a difference in skin tone between the individuals shown in the two panels (14 images), our analysis showed that the change was always from a darker to a lighter skin tone, perpetuating the social bias that a person must be White or light-skinned to make career progress.  Analysis of body types and occupations of the individuals revealed a preference of the model for the mesomorphs 
(athletic, solid, strong, not overweight or underweight). Endomorphs (lots of body fat or muscle) were rarely generated (8\%).  Images that depicted occupations in the areas of {\em Arts, Audio/Video Technology \& Communications} and {\em Business Management, Administration \& Finance} were predominately male (77\%). 

The "ground truth" of the attributes (prompted and unprompted) in this experiment were provided by four annotators. The Inter-Coder Reliability of each pair of annotators was calculated using the Cohen-Kappa score~\cite{Henry-etal-2017} and percent agreement based on an initial round of annotation on the first 40 images (10\% of the total number of images available). The Cohen-Kappa scores ranged from 0.64 (Good) to 1.0 (Very Good). The percent agreement scores were then used to guide the allocation of annotator pairs to re-evaluate images for which the Cohen-Kappa score was only in the "Good" range. Using this process, we eventually achieved a Cohen-Kappa score greater than 0.8)~\cite{Henry-etal-2017}, which is considered a robust  Inter-Coder Reliability for QCA~\cite{doi:10.1177/1609406919899220}.

\subsection{Ethical Intervention Experiments}

Based on our experimental results with the {\em Attribute List Prompt} described in Sect.~\ref{sec:two-panels}, in subsequent experiments with ethical intervention prompts, we adjusted the prompts shown in Figure~\ref{fig:prompts}.  We instructed the text-to-image model to generate a photograph-style image of a {\em single} person, facing forward in the generated image.  This then enabled us to automate the annotation process, using the tools described in Section~\ref{sec:automated-tools}.  We note that, given the list of 200 headlines, we could only prompt DALL-E 3 to process 5 headlines at a time, generating one image per headline, because it would hang when asked to generate more than five images at a time.  This resulted resulted in 40 iterations of DebiasPI process.  

Our experiments with the {\em Attribute Distribution Prompt} focused on the uniform distribution, asking the text-to-image model to produce outputs with equal representations of the attributes. As designed,  the DebiasPI process 
results in outputs with a perfect balance of gender or race. The attribute type "skin tone" was more difficult to handle, as we show below.

\begin{figure}[t]
    \begin{center}
       \includegraphics[height=140pt]{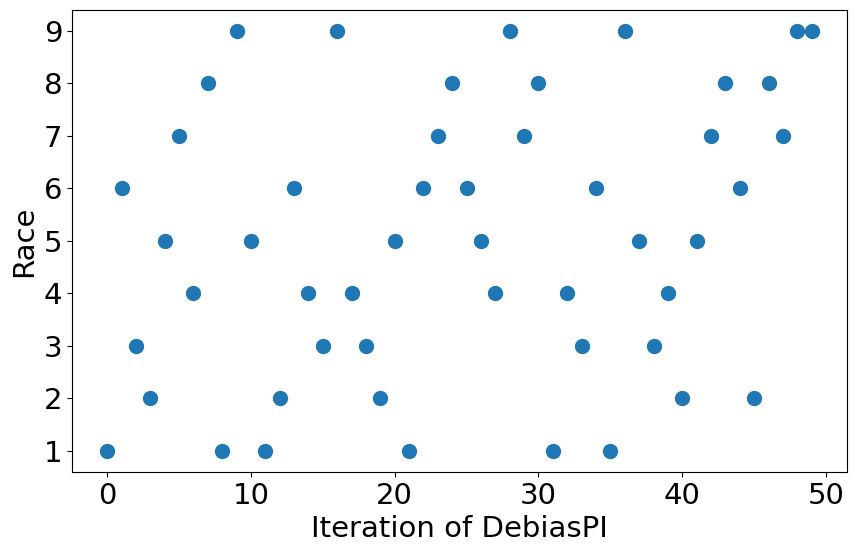} 
       \end{center}
    \caption{Race choices made by DebiasPI during a 50-image generation process.  Here, 6 Black individuals (race category 1) were frequently created by iteration 35, after the target is fulfilled the model does not generate any more black faces in subsequent iterations.}
    \label{fig:cycle-through-race-choices}
    \end{figure}

To illustrate how DebiasPI iterates through its attribute selections and yields a balanced outcome, we report on an 
experiment with 50 images created with a 9-race uniform distribution prompt (Fig.~\ref{fig:cycle-through-race-choices}). The plot shows how the selection of races is adjusted once the maximum (determined by the uniform distribution) for a specific race is reached. 


\noindent
{\bf Ablation Studies.} 
We then asked the question: "What if we ask for a uniform distribution but handicap DebiasPI by not allowing it to keep track of the choices at each step?"  The resulting "ablation studies" yield highly non-uniform distributions that show the text-to-image model's flawed interpretation of an equal representation of attributes.  This outcome underscores the challenges in achieving truly balanced outputs of the text-to-image model without direct intervention.  In the following, we describe the results of experiments with and without intervention on the attribute distributions of gender, race, age, and skin tone in detail.  


\noindent
{\bf Gender.} We used the attribute list gender=[male, female]. Prompting the text-to-image model with this list of gender options yielded 71\% males and 29\% females.  Prompting the model with an attribute list and desired uniform distribution, i.e., gender=[male, 50\%, female 50\%] for 200 images, yielded 56\% males and 44\% females, a moderately successful attempt at equal representation (in the absence of DebiasPI's explicit choice counting).   Interestingly, we found that when the model was alerted to ethical interventions with respect to race but not gender, it improved the representation of gender in its outputs.  Specifically, if race options were provided in the {\em Attribute List Prompt}, the resulting gender representation was 54\% male and 46\% female, and with a request for racial balancing, 52\% male and 48\% female.

\begin{figure}[t]
    \centering
    \includegraphics[width=0.48\linewidth]{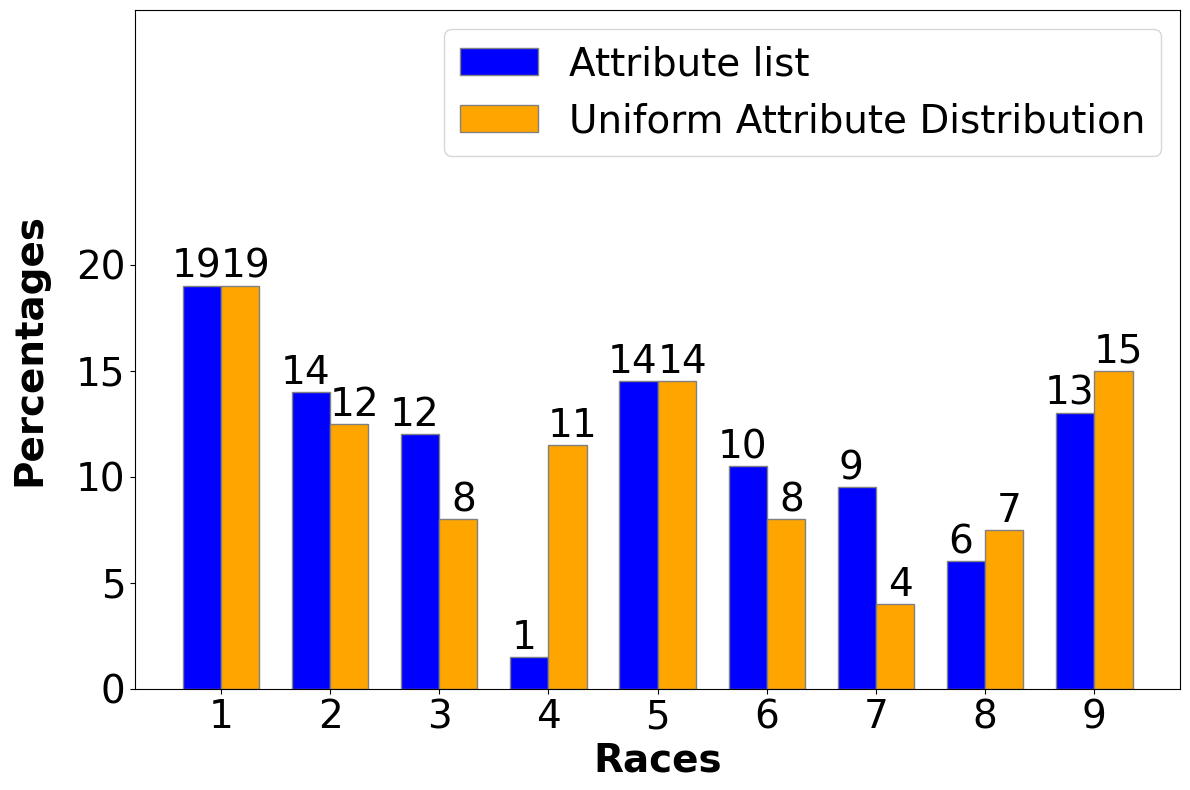} \hfill
    \includegraphics[width=0.48\linewidth]{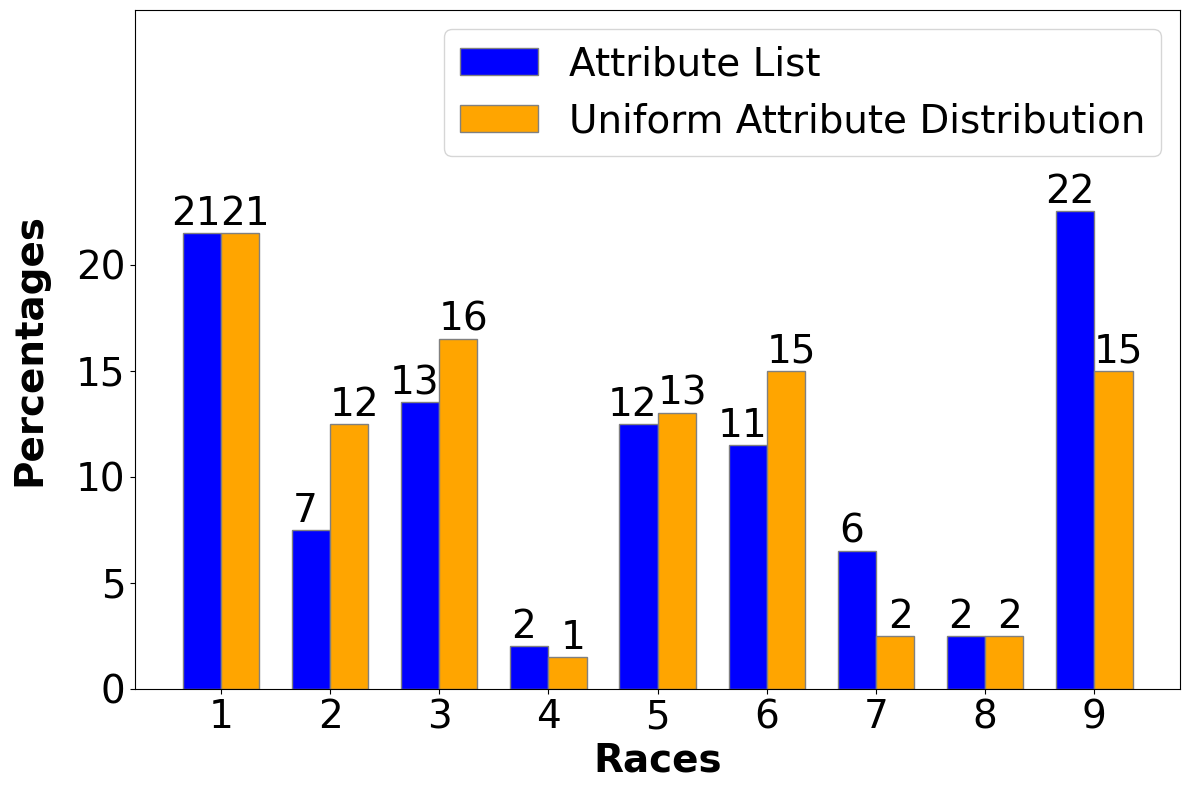}
    \caption{Ablation study: Prompting with attribute lists only (blue) and with attribute distributions (orange) without DebiasPI's choice counting.  Attribute lists are 9 race options (left) and 2 gender, 9 race options (right). The races are: Black, East Asian, Hispanic, Indigenous, Middle Eastern or North African, South Asian, Southeast Asian, Native Hawaiian and Other Pacific Islander, and White. The desired uniformity of the attribute distributions was not achieved by the text-to-image model in the absence of DebiasPI's choice counting.  EMD and JS-Div analysis shows that attempting to balance gender and race (right) results in greater bias compared to focusing solely on race (left)  
    (blue: EMD 0.04, JS-Div 0.02 (left) and EMD 0.06, JS-Div 0.06 (right)).  }
    \label{fig:race}
\end{figure}



\noindent
{\bf Race.} Next, we experimented with attribute lists race=[9 choices], and gender=[male, female], race=[9 choices]. The resulting race distributions are shown in Figure~\ref{fig:race} (again, in the absence of DebiasPI's explicit choice counting).
Since we had to prompt DALL-E~3 to generate five images at a time, for each dataset creation, the model was called to generate images 40 times to cover the 200 headlines. 
DALL-E~3 only has 5 chances to cover 9 races in each of the 40 trials, so if it draws race categories uniformly, the probability that a specific race will be chosen in each trial is 0.56 i.e., $p=(\binom{9}{5}-\binom{8}{5})/\binom{9}{5}$, meaning in expectation there will be $n \times p=40 \times 0.56\approx22$ images per race in the 200 generated images (or, 11$\pm\sigma$\% for each race where $\sigma=\sqrt{n \times p \times (1-p)}\approx3$). A simulation was run to confirm this,  
which showed that if the model attempted to draw race categories uniformly, then we should see between 8-14\% images for each category. When comparing our experimental results in Figure~\ref{fig:race} (blue versus orange distributions), we can see that DALL-E~3 indeed made an attempt to decrease over-representation of certain races while increasing the representation of under-represented ones. For example, Figure~\ref{fig:race}(right) shows a notable decrease of White
and an increase 
of Hispanic, Middle Eastern or North African, South Asian, and East Asian. 
However, Southeast Asian, Native Hawaiian and Other Pacific Islander and Indigenous populations remain under-represented. 
Analysis with JS-Div and EMD found that attempting to balance more than one set of attributes at the same time, here gender and race, results in greater bias compared to focusing solely on one type of attribute (race or gender).



\noindent
{\bf Age.} With the two ViT models (see Sec.~\ref{sec:automated-tools}), we obtained age estimates that placed the most common age of all generated images  in the age group [34-64] and very few individuals in age group [65+]. One model~\cite{nate_raw_2023}, however, estimated that the remaining individuals ($\approx$ 40\%) were in the age group [19-34], while the other model~\cite{dima806} estimated them to be in the age group [<19]. The majority of individuals in images inspected manually were young adults [19-34].   

\begin{figure}[t]
    \centering
    \hfill
   \includegraphics[height=120pt]{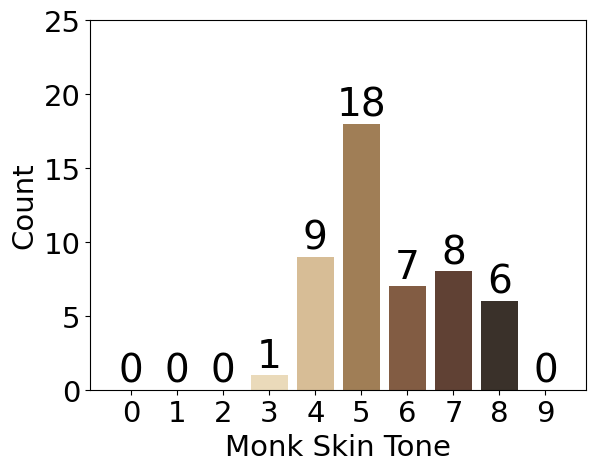} \hfill
      \includegraphics[height=120pt]{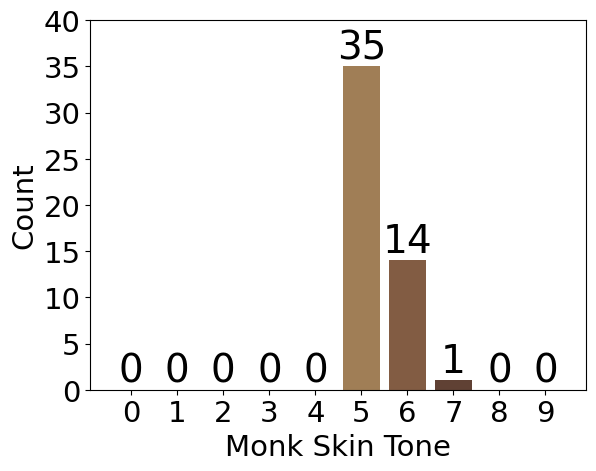} \hfill
    \caption{Skin tone distributions over 10-choice Monk scale for attribute list 5-skin-tones (left) and attribute list 9-race choices (right).}
    \label{fig:skin-color-debias}
    \end{figure}

\noindent
{\bf Skin Color.}
With the next set of experiments, we studied how we can instruct the AI model to generate images with a wide range of skin tones, now allowing DebiasPI to count attribute choices.  Results for iterating on DebiasPI 50 times with the 5-choice skin-tone list [dark tan, pale tan, purely black, warm brown, white fair] and the desired distribution uniform are shown in Figure~\ref{fig:skin-color-debias} (left).  If the attribute list is race and not skin tone, DebiasPI generates skin tone that is only medium brown (Fig.~\ref{fig:skin-color-debias} right). If DebiasPI has an input attribute distribution that is severely skewed toward light skin color, we found that DALL-E~3 was still not able to generate the palest four shades of the Monk scale. 
We acknowledge that these results were obtained in a relatively small experiment involving only 50 generated images but suggest that similar outcomes would be measured in larger experiments.

\noindent
{\bf Desired Non-uniform Distributions.}
Although our focus has been on desired distributions that are uniform,
we must note that DebiasPI can handle any desired input distribution.   For example, when we asked DebiasPI to generate 90\% female and 10\% male characters in a 50-image experiment, it produced the desired 45 female and 5 male character images, as verified by an external classifier of gender.   Another example involves an experiment with different desired race distributions (Table~\ref{table:debiaspi_on_race}).  Here, we first provided a 9-race attribute list without any distribution (Table~\ref{table:debiaspi_on_race}) row 1). The text-to-image model most frequently selected the category "Black," a result aligned with the first cycles of DebiasPI of the experiment shown in Figure~\ref{fig:cycle-through-race-choices}.   We then asked DebiasPI to produce a desired distribution that is uniform, which yielded in 5 or 6 images per race category (Table~\ref{table:debiaspi_on_race}, row 2).  Finally, we asked DebiasPI to produce 90\% "White" faces and distributes the remaining 10\% selections uniformly among the other 8 races. This resulted in 40 images in the "White" race category, and one or two images in the other categories (Table~\ref{table:debiaspi_on_race}, row 3).

\begin{table}[thb]
\caption{Race Distribution Obtained by DebiasPI for 50 images, without choice counting (row 1), with choice counting and a desired uniform distribution (row 2), and with choice counting and a desired non-uniform distribution (row 3).  
DebiasPI reaches the set distribution targets exactly (rows~2 and~3). Abbreviations: ME/NA (Middle Eastern or North African), NH/PI (Native Hawaiian or Pacific Islander), SE (Southeast).}
\centering
\renewcommand\cellalign{lc}
\small
\setcellgapes{3pt}\makegapedcells
\begin{tabular}{c||c|c|c|c|c|c|c|c|c} \hline
\makecell{Desired\\Distribution} & \makecell{Black} & \makecell{East\\Asian} & \makecell{Hispanic} & \makecell{Indigenous} & \makecell{ME/NA} & \makecell{South\\Asian} & \makecell{NH/PI} & \makecell{SE\\Asian} & \makecell{White} \\ \hline
\makecell{None} & 10 & 5 & 7 & 4 & 7 & 7 & 1 & 4 & 5 \\ \hline
\makecell{Uniform}  & 6 & 5 & 5 & 6 & 6 & 6 & 5 & 5 & 6 \\ \hline
\makecell{Non-uniform}  & 1 & 1 & 2 & 1 & 1 & 1 & 2 & 1 & 40 \\ \hline
\end{tabular}
\label{table:debiaspi_on_race}
\end{table}


\section{Conclusions}


We proposed DebiasPI, an inference-time framework for robust attribute distribution control.  With DebiasPI, a user can generate a series of images with attributes aligned to a target distribution, such as uniform distribution for fairness or a distribution that stresses specific traits. We envision, as a use case, a newsroom editor who might want to select among a diverse set of images of athletes.  Our experiments show that DebiasPI is successful in generating images for representation of race and gender according to the desired attribute distribution.  

Limitations of our work include the relatively small numbers of experimental data, the dependence of DebiasPI on the text-to-image model's ability to generate certain attributes (for example, skin tone), and the challenge that the model's internal beliefs or the external classifier's attribute analysis may not be entirely reliable, complicating DebiasPI's control over outputs.  

The datasets we here publish may serve as benchmark comparisons by others in future work.  Additional experiments with body-type, age, profession, and not-yet-explored attributes like affect would be interesting. Future work will also study ethical intervention when the text-to-image model is challenged with abstract concepts in the text to be visualized.


%
%

\section*{Acknowledgements}
This work was supported in part by the U.S.\ National Science Foundation under grant No.\  1838193.  
Any opinions, findings, and conclusions or recommendations expressed in this material are those of the authors and do not necessarily reflect the views of the National Science Foundation. The authors thank the members of the Artificial Intelligence and Emerging Media (AIEM) Research Group at Boston University, {\tt https://sites.bu.edu/aiem}.

\bibliographystyle{splncs04}

\end{document}